\documentclass[11pt,a4paper]{article}
\usepackage[hyperref]{emnlp2018}
\usepackage{times}
\usepackage{latexsym}
\usepackage{graphicx}
\usepackage{amsmath}
\usepackage{mathptmx}
\usepackage{amssymb}
\usepackage{verbatim}
\usepackage{url}

\aclfinalcopy 
\def\aclpaperid{658} 

\title{An Instance Transfer based Approach Using Enhanced Recurrent Neural Network for Domain Named Entity Recognition}

\author{Lin Li \\
School of Computer Sci \& Technol, \\ Wuhan University of Technology, China\\
{\tt cathylilin@whut.edu.cn} \\\And
Yueqing Sun \\
School of Computer Sci \& Technol, \\ Wuhan University of Technology, China\\
{\tt yqsuan@whut.edu.cn} \\
}

\date{}

\begin{document}
\maketitle
\begin{abstract}
Recently, neural networks have shown promising results for named entity recognition (NER), which needs a number of labeled data to for model training. When meeting a new domain (target domain) for NER, there is no or a few labeled data, which makes domain NER much more difficult. As NER has been researched for a long time, some similar domain already has well labelled data (source domain). Therefore, in this paper, we focus on domain NER by studying how to utilize the labelled data from such similar source domain for the new target domain. We design a kernel function based instance transfer strategy by getting similar labelled sentences from a source domain. Moreover, we propose an enhanced recurrent neural network (ERNN) by adding an additional layer that combines the source domain labelled data into traditional RNN structure. Comprehensive experiments are conducted on two datasets. The comparison results among HMM, CRF and RNN show that RNN performs bette than others. When there is no labelled data in domain target, compared to directly using the source domain labelled data without selecting transferred instances, our enhanced RNN approach gets improvement from 0.8052 to 0.9328 in terms of F1 measure.
\end{abstract}

\section{Introduction}\label{intro}
In recent years, Web data and knowledge management attracts the interests from industry and research fields. There are various promising applications, such as intelligent recommendation, machine Question \& Answer, knowledge graph and so on. Named entity recognition is a fundamental and very important step in the automatic information extraction. A recognition method with high quality can directly improve the follow-up processing results of Web data management products. NER research shows great successes in various domains and becomes a hot topic~\cite{6,7}, such as social media~\cite{1,2}, language texts~\cite{3}, biomedicine~\cite{4,5}, and so on. As we know, texts of different domains may vary from features, writing styles and structures. Domain NER meets a challenge that annotating data for new domains is labor intensive.

For domain NER, a new domain (target domain) has no or a few labelled data. However, it is natural to think that if some similar domain (source domain) with enough lablled data already exists, it is possible to borrow some from this similar domain. Domain adaptation targets at transferring the source domain knowledge to the target domain~\cite{da}) , which is a effective way to solve the problem of labelling large amount of data on new corpus or domains. In other words, if a NER modle is trained well for some fixed source domain, it is interesting to study how to deploy them across one or more different target domains. 

In this paper, we focus on domain NER for the politics text domain in Chinese high schools to support an automatic question and answer system (Q\&A) that will take the national college entrance examination (NCEE) in the future. However there is no public politics text corpus used by Chinese high school for NER task. Moreover, there is no labeled data. We notice that People's Daily corpus is free for public download and similar to politics text. Therefore, we propose an instance transfer based approach for domain NER with enhanced Recurrent Neural Network (RNN). First, we design an instance transfer strategy to extract similar sentences from the People's Daily corpus (source domain). Here, instances mean labelled data and politics text is our target domain. Then, recurrent neural network (RNN) model can be trained based on the transferred instances. Moreover, we improve the traditional RNN model by enhancing its activation function and structure. Finally, an instance transfer enhanced RNN (ERNN) model is proposed to do NER for politics text target domain, which is trained based on the transferred instances from similar source domain (People's daily corpus). Compared with the traditional RNN model, experimental results show that our ERNN with the instance transfer strategy can get improvement from 80.52\% to 93.28\% in terms of F1 measure.


In addition, we consider other situation where a small number of labelled data is available to further investigate the performance of our proposed approach. Since the politics text as target domain data is needed to obtain, we collect texts from high school books and relevant websites and manually label a very small set of them, making labeled data much less than the unlabeled one. Experimental results show that labeling data for target domain is quite useful, reaching at 92.13 in terms of F1 measure. However, our instance transfer based enhanced RNN can further improve the F1 value to 93.81. Finally, we adopt the co-training approach using our proposed ERNN model and CRF by taking advantage of large unannotated target domain data, which get the F1 value of 94.02.


The rest of the paper is organized as follows. In Section~\ref{related}, we introduce related work on NER (especially on domain NER), recurrent neural network and transfer learning. Section~\ref{approach} elaborates our approach including the instance transfer strategy, our proposed ERNN. Experiments and results are given in Section~\ref{exp}, then we make the conclusion and future work in Section~i\ref{conclusion}.

\section{Related Work}\label{related}
As the virtual step in information extraction and a shared task of CoNLL-2003~\cite{8}, NER has been widely studied nowadays and most are about domain NER. Zhang et al.~\shortcite{9} employ conditional random fields (CRF) and structure support vector machines (SSVMs) to do chemical entity mention recognition in patients and chemical passage detection. Chen et al.~\shortcite{10} propose novel active learning algorithms to significantly save annotation cost for clinical NER task. Besides, considering the condition that available labeled data is not at hand sometimes, Brooke et al.~\shortcite{11} present a NER system for tagging fiction. To avoid the lack of annotated data, they bootstrap a model from term clusters and leverage multiple instances of the same name in a text instead. However it needs to pass through the corpus to build a feature vector corresponding to all the contexts in one document. Moreover, a large number of researches about domain NER are on biomedical fields~\cite{12,13}, and the typical methods are CRF~\cite{14,15}, and other supervised learning approaches~\cite{16}.

As neural networks have shown promising results in domain NER~\cite{13} , such as RNN, LSTM, and so on. Tomori et al.~\shortcite{17} use deep neural networks for referring to the real world (i.e. game states) to improve Japanese chess NER. Zeng and Sun et al.~\shortcite{18} combine the bidirectional long short-term memory (LSTM) and CRF to automatically explore words and characters level features. However, transfer learning and semi-supervised learning, like co-training, which are effective ways for the situation when train data is much less than unannotated data. 

When considering how to deal with the situation of lack of labelled data, Pen and Yang~\shortcite{24} have categorized and reviewed the reserch progress on transfer learning for classification, regression, and clustering problems. Similar to the work of Qu et al.~\shortcite{21} in domain adaption, in this paper we study a instance transfer strategy, which is hot today but rarely used in NER~\cite{22,23}, to make use of out-of-domain data (source domain). 

Co-training is also an effecitve way to use unlabelled data in supervised learning task, Li and Huang et al.~\shortcite{19} use a bilingual co-training and maximum entropy model to carry out English and Chinese NER. Munkhdalai et al.~\shortcite{20} modify the original co-training to cover knowledge from unlabeled data to recognize bio named entities in text. Besides, clustering whould be another research direction for this problem~\cite{YangW1,YangW2,YangW3}. Our interest is domain NER for Chinese high school news texts. In this paper we also put our proposed ERNN into a co-training approach to see how much improvement can be obtained through large amounts of unlabeled data.


The contributions and differences of our work from others are listed as follow:
\begin{enumerate}
\item We design a instance transfer strategy by selecting similar sentences from a source domain.
\item We add an additional layer to the traditional RNN structure by coupling knowledge from the transferred similar instances.
\item We conduct experiments on two datasets. No matter there is no labelled data or no in target domain, our proposed approach can improve the quality of NER.
\end{enumerate}

\section{Our Approach}\label{approach}
\subsection{Problem Description}
\label{approach:1}
Manually labeling data is time consuming in most machine learning methods. However, there are some other domain data with well-labeled data. As Pen and Yang have declared~\shortcite{24}, in such cases, knowledge transfer, if done successfully, would greatly improve the performance of learning without taking lots of efforts to label data. Our interest is Chinese high school politics text NER to support automatic Question \& Answer application. There is no labelled data at hand for this new domain, but we find that some public NER data, like People's Daily corpus, is similar to it. The similar domain with well-labeled data is called source domain, while the new domain without enough labelled data is called target domain. In this paper, we propose an instance transfer enhanced RNN model for domain NER by leveraging labelled data from a source domain. In Section~\ref{approach:2}, we will introduce an our instance transfer strategy that will select similar labeled data (sentences) from a source domain. In Section~\ref{approach:3}, we will discuss how to improve the traditional RNN model by adding those selected data into the network structure of RNN.

\subsection{Our Instance Transfer Strategy}
\label{approach:2}
We design a instance transfer strategy where the target domain is Chinese high school politics texts, and People's Daily is the source domain. Since texts from different domains may vary from features, writing styles and structures, it is not suitable to use all labelled data from the source domain. In this paper we first consider two functions to compute sentence similarity between source domain and target domain. The two functions are Gaussian radial basis function (RBF) and polynomial kernel function which are described as following:
\begin{itemize}
\item Gaussian RBF kernel:
\begin{equation}\label{RBF}
K(\vec{x},\vec{z})=exp(-\frac{(\left \lVert \vec{x}-\vec{z} \right \rVert)^2}{2\sigma^2})
\end{equation}
\item polynomial kernel function:
\begin{equation}\label{polynomial}
K(\vec{x},\vec{z})=(\vec{x}\bullet \vec{z}+1)^{d}
\end{equation}
\end{itemize}

Here the $\vec{x}$ stands for the sentence vector in source domain, while $\vec{z}$ represents the mean of target domain corpus matrix. And we set $\sigma$=1 and d=2. We use these two functions to evaluate the sentence similarity between source domain and target domain and sort source domain sentences according to their similarity scores. 

Then two transfer strategies are given to select top similar sentences:
\begin{enumerate}
\item Directly regard the top n sentences as target domain training data.
\item Repeat the nth sentence k times in source domain where the value of k depends on n. The original source corpus has been enlarged.
\end{enumerate}

\subsection{Our ERNN Model}
\label{approach:3}
Through our instance transfer strategy, target domain borrows labelled data from source domain and top similar sentences have been assigned more weights than others. Therefore, the next step of our approach is to enhance RNN model by utilizing those labelled data. RNN is designed to solve the problems involved in time and sequence~\cite{25}. Not only the output of moment $t$ is influenced by that of previous moment, but also the nodes between hidden layer are all connected to each other. In this paper, we propose an enhanced RNN by modifying both its activation function and structure to do domain NER task.

\subsubsection{Activation Function}
\label{approach:3:1}
As described in~\cite{31}, a deep neural network is an ensemble of vectors and matrices that stand for bias and weight values, and nonlinear activation function. 
A suitable activation function means much for a deep neural network. Its changes can speed up model training~\cite{32} and enhance stability~\cite{33}, etc. However, for a given domain, the best nonlinear function remains unknown, even though many rectifier-type nonlinear functions have been proposed as activation functions~\cite{31}. The performance of a same activation function may be widely different when applying it to different tasks. For NER task, compared to other frequently used activation functions such as ReLu, PReLu, tanh and so on, the sigmoid function shows best in our preliminary experiments. Its definition is given in Eq.~\ref{sigmod}.
\begin{equation}\label{sigmod}
F(x) = \frac{1}{1+e^{-x}}
\end{equation}

\begin{figure}[hbt]
\includegraphics[scale=0.45]{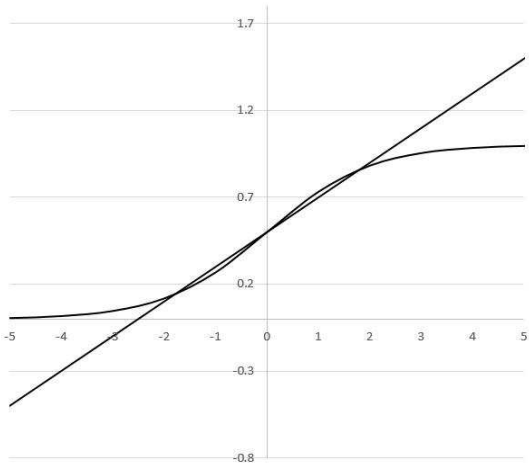}
\caption{Sigmoid and its linear approximation function}
\label{function}
\end{figure}

However, the model training process involves the derivation, while the derivative of sigmoid function tends to zero when the argument x approaches infinity (which called the saturation phenomenon) both at its left and right. This may make it harder to train model. Even so, sigmoid is most similar to the reflex mechanism on the biological neuron level and the interval of its output is always between 0 and 1, which can represent the prediction probability of a label. In our experiments, its linear approximation function expressed by Eq.~\ref{linearsigmod}, shows a good performance and a is set to be 0.2 and b is 0.5 A more intuitive depiction is shown in Figure \ref{function}.
\begin{equation}\label{linearsigmod}
L(x) = a\times x+b
\end{equation}

Focusing on our NER task, a new activation function is proposed by combining the two above functions (shown in Eq.~\ref{combined}). Here parameters $\alpha$ and $\beta$ are coefficients and they are determined by experimental performances.
\begin{equation}\label{combined}
A(x) = \alpha \times F(x)+\beta \times L(x)
\end{equation}

The new activation function has some advantages compared to its two original functions: sigmoid and linear approximation. It ameliorates the former's saturation phenomenon on the one hand and smooths the latter on the other hand. In our experiments, the results using the proposed activation function are better than that using sigmoid or its linear approximation.

\subsubsection{Model Structure}
\label{approach:3:2}
In order to take full advantage of source domain data, we also improve RNN's structure and propose an instance transfer enhanced RNN (ERNN), which can input the source domain sentences into RNN. Our ERNN is modified based on the Elman-type described in~\cite{26}. 

\begin{figure}[htb]
\includegraphics[scale=0.45]{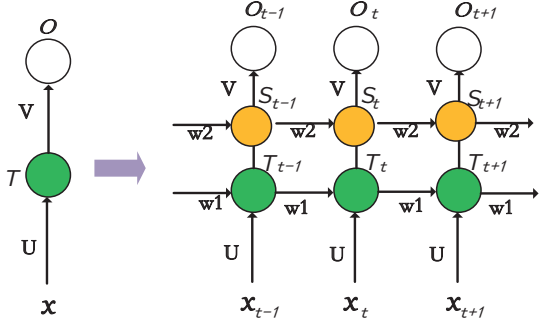}
\caption{The RNN structure before/after modified}
\label{fig:detailRNN}
\end{figure}

\begin{figure}[htb]
\includegraphics[scale=0.30]{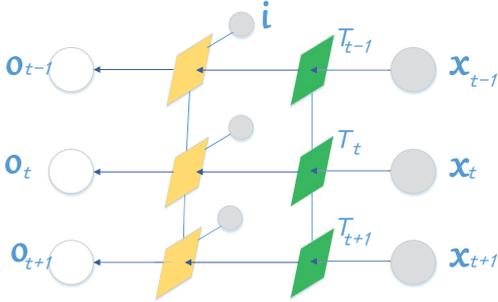}
\caption{The overview of our ERNN}
\label{fig:overviewRNN}
\end{figure}

The new structure we modified is depicted as Figure~\ref{fig:detailRNN} and Figure~\ref{fig:overviewRNN}. In Figure~\ref{fig:detailRNN}, the left shows for the original structure and the right describes our modified recurrent neural network when unfolding its structure in time of the computation involved in its forward computation. $T$ nodes represent the hidden layer, meanwhile $S$ indicates the confluent layer. The source domain data combines with the output of the hidden layer $T$. The overview of the structure is shown in Figure~\ref{fig:overviewRNN} that depicts the domain source data input $i$ compared to Figure~\ref{fig:detailRNN}. As shown in Figure~\ref{fig:detailRNN}, $\vec{U}$, $\vec{V}$, $\vec{w}$, $\vec{w_1}$, $\vec{w_2}$ are weights and parameters. The output of the $S_t$ is computed by Eq.~\ref{st}. In output layer, we use softmax to get the prediction results i.e. $o(t$), as shown in Eq.~\ref{ot}.

\begin{equation}\label{st}
S_t=F(Tt\cdot \vec{W}+\emph{i}\cdot \vec{w_2}+\vec{b_0})
\end{equation}

\begin{equation}\label{ot}
O(t)=softmax(S_t\cdot \vec{V}+\vec{b_1})
\end{equation}

Here $F$ is the activation function, \emph{i} stands for the source domain data, as marked in Figure\ref{fig:overviewRNN}. $\vec{b_0}$ and $\vec{b_1}$ are bias vectors. In this way we achieve the reuse of the valuable source domain corpus.

\section{Experiments}
\label{exp}
We conduct two sets of experiments on two totally different datasets for different purposes. One is NER experiments among three popular models, i.e., HMM, CRF and RNN, which aims to explore the classification quality with different training data sizes. The other is domain NER using our ERNN with instance transfer.

\subsection{Data Description}
\label{exp:1}
There are two datasets used in our experiments. For NER experiment among three popular models, we mainly focus the changes of the classification performance with different amount of training data. We conduct the experiment on a standard corpus named ATIS~\cite{26}. It contains 127 classes and uses the in/out/begin (IOB) representation. A sentence in ATIS can be expressed as in Table~\ref{ATIS}.

The other dataset is collected from Chinese high school politics textbooks and websites, which is the target domain in this paper. The source domain corpus is People's Daily 1998 that is open online. In addition, due to lack of label information, we set 11 classes for target domain based on People's Daily 1998, including person names, regular time words, single time words such as those means "night", proprietary organization and company name. For example, Chinese characters are "Xinhua News Agency" and "the United Front Department", and special region name, noun and etc.

\begin{table*}[t!]
\begin{center}
\begin{tabular}{|l|l|l|l|l|l|}
\hline
\bf Sentence & flight & from & memphis & to & tacoma \\
\bf Label & O & O & B-fromloc & O & B-toloc \\
& & & .city\_name & & .city\_name \\
\hline
\bf Sentence & limousine & service & at & logan & airport\\
\bf Label & B-transport\_type & O & O & B-toloc.airport\_name & I-airport\_name\\
\hline
\end{tabular}
\end{center}
\caption{Sentences in ATIS} \label{ATIS}
\end{table*}

\subsection{Data Pre-processing}
\label{exp:2}
Specifically, for NER experiment, ATIS contains 3983 training sentences and 893 testing sentences. In order to find out how it works when the labeled data is less than the unlabeled one, we reorganized the training set. Based on ATIS original train/test proportion (3983/893), we randomly select 20\%, 40\%, 60\%, 80\% and 100\% of total train data (3983 sentences) as our training sets to train two traditional statistical models(Hidden Markov Model and CRF) and RNN model and one RNN model. For domain NER experiment, we first pick the most common words from target domain data as our dictionary. By counting the times a word has appeared in this corpus, we select top 13,450 words as our dictionary and other words are marked as 'unknown' in a sentence. After this, we clean the data based on a rule that is if a sentence meets any of the following two conditions, it would be regarded as noisy data and abandoned.
\begin{itemize}
\item The length is less than 3
\item Too many 'unknown' tags(more than 50\%)
\end{itemize}

After preprocessing, the People's Daily sentences have been reduced from 19484 to 4818, meanwhile the number of politics text sentences is also reduced from 82700 to 15584. Since there is no annotated data of the latter, we manually mark 3043 sentences. Table~\ref{datasets} shows the statistics of the labeled and unlabeled data size of both datasets. 
\begin{table}[htb]
\begin{center}
\begin{tabular}{|l|l|l|l|l|}
\hline
{\bfseries Corpus-} & {\bfseries Labeled}& {\bfseries Train}&{\bfseries Test} & {\bfseries Unla-} \\
{\bfseries Domain} & {\bfseries Data}& {\bfseries Set}&{\bfseries Set} & {\bfseries beled} \\
\hline
& & & & \\
Source & 4818 & 3855 & 963 & 0 \\
\hline
& 3043(manu- & & & \\
Target & ally labeled) &2035 & 1008 & 13549 \\
\hline
\end{tabular}
\end{center}
\caption{\label{datasets}The statistcs of domain and target datasets}
\end{table}

For People's Daily, we use 80\%/20\% of the total 4818 sentences as our train/test data sets. For our target domain data in severe lack of tagged data, we use manually labeled 2035 sentences and tag other 1008 sentences from the large unannotated set as the test set.

\subsection{Evaluation and Setup}
\label{exp:3}
In this paper, all the experimental results were evaluated by the popular evaluation measures, i.e., precision (P), recall (R) and F1-score. 
In addition, for supervised learning experiment, we used K-fold cross validations. K is different based on the different size of train set. For example, when using 20\% of the original train set, K is 5, and when using 40\% and 60\%, K is 3.

We also do some preliminary experiments to set the ERNN parameters. For example, the word embedding we use is trained by target domain texts via word2vec, a Google open source project which brings about 14.1\% performance improvement. Besides, we set the ERNN context window size equal to 1, which performs better than those 5, 7 and 9 (probably because there are too many non-entity labels in a sentence). For instance transfer, since the RBF similarity function achieves a better result, we use it to enlarge the source domain corpus.

\subsection{ NER Experimental Results of Three Popular NER Models}
\label{exp:4}
\begin{figure}
\includegraphics[scale=0.65]{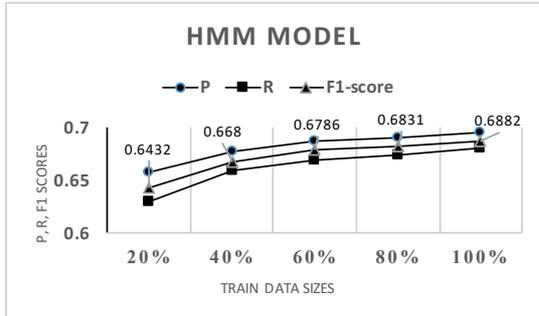}
\caption{Results of the HMM model}\label{dataszie1}
\end{figure}

\begin{figure}
\includegraphics[scale=0.65]{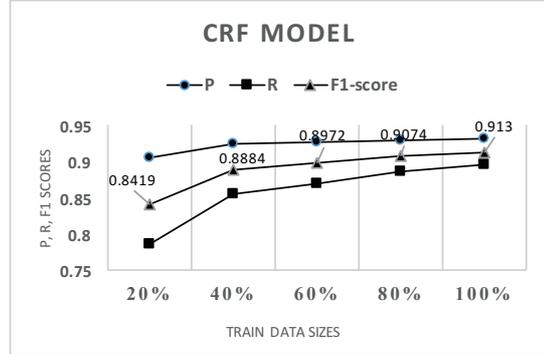}
\caption{Results of the CRF model}\label{dataszie2}
\end{figure}

\begin{figure}
\includegraphics[scale=0.65]{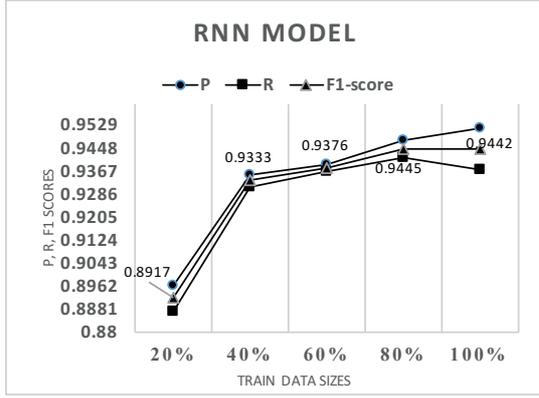}
\caption{Results of RNN model}\label{dataszie3}
\end{figure}

In this part, we compare the NER performance of a current popular deep learning model, i.e., RNN with two traditional models, i.e., HMM and CRF. And we also mainly explore the NER performance with different training data sizes. We conduct several experiments using ATIS's total training data's 20\%, 40\%, 60\%, 80\% and 100\% to train the three popular NER models. The results are shown in Figure~\ref{dataszie1}, Figure~\ref{dataszie2} and Figure~\ref{dataszie3}. 

From Figure~\ref{dataszie1}, Figure~\ref{dataszie2} and Figure~\ref{dataszie3} we can see how the three models behave when being trained on different data sizes. Overall, the P, R and F1 of these three models are gradually rising when using more training data. However, the performances are not growing linearly when the data size gets larger. While using more and more data to train a model, the degree of improvement becomes smaller. Actually, when the training data size increases from 20\% to 40\%, the three models generally achieve a highest improvement (about 4\% improvement in terms of F1 measure). After that, increasing the training data size brings little benefits. 

This indicates that when meeting a new domain for NER, labelled data will actually improve performance. But manully labelling data is labor intensive. Putting more labor on labelling data could not bring us larger improvement considering the time and the labor we need. Therefore, in this paper, we propose an instance transfer strategy by transferring labelled data from other well labelled domain, which is a effective way to solve the problem of lack of labelled data in a new domain.

\subsection{Domain NER Experimental Results}
\label{exp:6}
\subsubsection{Instance Transfer}
\label{exp:6:1}
We have designed the two transfer learning strategies introduced in Section~\ref{approach:2}. One is directly using the RBF top n (here we set n=800) similar sentences from source domain as new additional target domain training data. The other is increasing the source corpus by Eq.~\ref{enlarge}, i.e. the nth most similar sentences repeat k times. Our preliminary results show that the latter is better. Therefore, due to page limit, we just discuss the latter in this paper. Finally, 41500 sentences are obtained as labelled data. We still keep the 80\%/20\% train/test proportion when using the enlarged training data to train our ERNN model. The parameters are given in Equation~\ref{enlarge}.
\begin{equation}\label{enlarge}
\begin{cases}
k=80 \quad if \quad 1\leq or \le n\leq or \le 250 \\
k=50\quad if \quad 250\leq n\leq or \le 500 \\
k=30\quad if \quad 500\leq n\leq or \le 800 \\
k=1\quad if \quad 800\leq n
\end{cases}
\end{equation}

Since in the above experiment, RNN show better performance than HMM and CRF, RNN related baselines and our approach are listed as follows(Here 'IT' means 'Instance Transfer'). As discussed in Section~\ref{related}, most of current researches that study deep neural network for NER work on trying different variants of deep learning models by assuming that a number of training data is ready. However, we consider two situations. One is that there is no labelled data in target domain, i.e., RNN\_D\_IT and ERNN\_IT. The other is there is a few labelled data in target domain, i.e., RNN\_L, RNN\_L\_D\_IT and ERNN\_L\_IT. Our problem context is quite different from most of current work.
\begin{enumerate}
\item RNN\_D\_IT: train a traditional RNN model (left part in Figure~\ref{fig:detailRNN}) directly with source domain. This training process is intuitive and any NER model can be applied like this. 
\item ERNN\_IT: train our ERNN model (Figure~~\ref{fig:overviewRNN}) with selected instances from source domain by our instance transfer strategy (Eq.~\ref{enlarge}).
\item RNN\_L: train a traditional RNN model (left part in Figure~\ref{fig:detailRNN})with a few lablled data from target domain. This is a traditional training process for supervised learning and more results are discussed in Section~\ref{exp:4}. 
\item RNN\_L\_D\_IT: train a traditional RNN model (left part in Figure~\ref{fig:detailRNN}) with a few lablled data from target domain and all instances in source domain. The problem context is the same as that of Qu et al.~\shortcite{21}, but RNN is used in our paper instead of CRF.
\item ERNN\_L\_IT: train our ERNN model (Figure~~\ref{fig:overviewRNN}) with a few lablled data from target domain and transferred instances in source domain by our instance transfer strategy (Eq.~\ref{enlarge}).
\end{enumerate}

For the situation that there is no lablled data in target domain, the experimental results are reported in Table~\ref{iresults}. The results of RNN\_D\_IT tell us that directly using all the labelled data from source domain directly is not much satisfactory. With the help of our instance transfer strategy, similar sentences are selected and repeatedly used as training data. The F1 measure of our ERNN\_IT reaches at 93.28\%, with the improvement about 15.84\%.
\begin{table}[htbp]
\begin{center}
\begin{tabular}{|l|l|l|l|}
\hline
{\bfseries Experiments} & {\bfseries P(\%)}& {\bfseries R(\%)}&{\bfseries F1(\%)} \\
\hline
RNN\_D\_IT & 76.11 & 85.48 & 80.52 \\
ERNN\_IT & 92.67 & 93.90 & {\bfseries 93.28} \\
\hline
\end{tabular}
\end{center}
\caption{\label{iresults}P, R and F1 scores without labelled data}
\end{table}

In addition, the other situation for domain NER is considered where a few labelled data is available. Therefore, experiments are conducted by adding our manually labelled data into RNN\_D\_IT and ERNN\_IT. Experimental results are show in~\ref{iresults2}. Traditional RNN model gets the F1 value of 92.13, which further indicates that manually labelled data for target domain can largely improve the quality of domain NER. Under this situation. our ERNN\_L\_IT can still obtain the F1 value of 93.81, and it benefits from our designed instance transfer strategy and adding a particular layer to RNN. The F1 value of RNN\_L\_D\_T is 93.06, which says that source domain can help target domain NER. Transfer strategy should be studied. The results in Table~\ref{iresults} and Table~\ref{iresults} tell us that transfer learning is an effective way to leverage the lablled data from source domain corpus. 
\begin{table}[htbp]
\begin{center}
\begin{tabular}{|l|l|l|l|}
\hline
{\bfseries Experiments} & {\bfseries P(\%)}& {\bfseries R(\%)}&{\bfseries F1(\%)} \\
\hline
RNN\_L & 91.84 & 92.42 & 92.13 \\
RNN\_L\_D\_T & 94.57 & 91.60 & 93.06 \\
ERNN\_L\_IT & 93.16 & 94.47 & {\bfseries 93.81} \\
\hline
\end{tabular}
\end{center}
\caption{\label{iresults2}P, R and F1 scores with a few labelled data}
\end{table}

\subsubsection{Co-training}
\label{exp:6:2}
The experiments in instance transfer part aim to utilize the source domain data to raise the recognition performances. Along with instance transfer learning methods, we also consider taking advantages of unannotated data since we have much of them at hand. Specifically, we conduct experiments by co-training our ERNN and a statistical probability model i.e. CRF. Co-training is kind of a semi-supervised strategy that is firstly proposed by Blum and Mitchell~\shortcite{28} in 1998 and the conditional independence of the data views is declared as a required criterion. However, Abney~\shortcite{29} shows that the independence assumption can be relaxed, which means co-training is still effective under a weaker independence assumption. In this paper, we explore the effect of co-training with our ERNN and adopt it to leverage the large unannotated in-domain data.

The initial training data size is 2035 sentences here, while the unannotated data is 12541 sentences. We select top 800 high confidence level sentences in each iteration and after iterated 10 times, all 12541 sentences get labeled. Table~\ref{cresults} shows the results of each model in each iteration.
\begin{table}
\begin{center}
\begin{tabular}{|l|l|l|l|l|l|l|}
\hline
\multicolumn{3} {c} {\bf CRF (\%)} & \multicolumn{3} {c} {\bf ERNN (\%)} \\
\cline{1-6}
{\bfseries P} & {\bfseries R} & {\bfseries F1}& {\bfseries P} & {\bfseries R} & {\bfseries F1} \\
\hline
94.77 & 84.57 & 89.38 & 93.16 & 94.47 & 93.81 \\

95.47 & 84.95 & 89.90 & 92.50 & 94.96 & 93.71 \\
95.57 & 86.25 & 90.67 & 93.84 & 93.39 & 93.61 \\
96.18 & 86.49 & 91.08 & 93.92 & 93.88 & 93.90 \\
96.02 & 86.86 & 91.21 & 93.04 & 94.81 & 93.92 \\
96.70 & 86.57 & 91.36 & 95.04 & 93.01 & {\bfseries94.02} \\
96.61 & 87.12 & 91.62 & 94.23 & 93.75 & 93.99 \\
96.89 & 87.31 & 91.85 & 94.51 & 93.17 & 93.83 \\
96.65 & 87.77 & 92.00 & 94.31 & 93.70 & 94.00 \\
96.92 & 88.09 & 92.30 & 95.16 & 91.68 & 93.39 \\
\hline
\end{tabular}
\end{center}
\caption{\label{cresults}P, R and F1 score of ERNN and CRF in each iteration of co-training}
\end{table}


For the co-training experiments, large unannotated data also brings rise to the quality of NER, but not as much as that of instance transfer experiments. Due to the CRF's low starting point, the ERNN F1 score goes down after reaching 94.02 in terms of F1 measure.

\section{Conclusions and Future Work}
\label{conclusion}
Domain NER is important and useful in various applications. In this paper, we study instance transfer and RNN to improve the quality of domain NER. We leverage the source domain data by proposing an instance transfer enhanced RNN called ERNN. In addition, we adopt co-training strategy to leverage the large unannotated in-domain data to further improve the recognition performances. 
In the future, 
we are going to make a further exploration to automatic QA system, relationship extraction between entities.

\bibliography{emnlp2018sun}
\bibliographystyle{acl_natbib_nourl}

\end{document}